\documentclass[conference]{IEEEtran}
\usepackage[utf8]{inputenc}
\usepackage{cite}

\usepackage{booktabs}
\usepackage{multirow}
\usepackage{array}

\usepackage[colorlinks=true,linkcolor=blue,citecolor=red,urlcolor=blue]{hyperref}
\usepackage{amsmath,amssymb,amsfonts}
\usepackage{algorithmic}
\usepackage{graphicx}
\usepackage{textcomp}
\usepackage{xcolor}
\usepackage{booktabs}
\usepackage{tikz}
\usetikzlibrary{calc}
\usetikzlibrary{positioning,shapes,calc}
\usetikzlibrary{arrows.meta,positioning,calc,fit}

\usepackage{tikz}
\usetikzlibrary{arrows.meta,positioning,calc}

\definecolor{cInp}{HTML}{E3F2FD}   
\definecolor{cFeat}{HTML}{E8F5E9}  
\definecolor{cProj}{HTML}{FFF3E0}  
\definecolor{cNorm}{HTML}{F3E5F5}  
\definecolor{cGate}{HTML}{E1F5FE}  
\definecolor{cSSM}{HTML}{FFFDE7}   

\tikzset{
  >={Latex[length=2.5mm]},
  every path/.style={line width=0.5pt},
  qbox/.style={draw, rounded corners, thick, inner sep=3pt, minimum height=7mm, align=center},
  qop/.style ={draw, rounded corners, thick, inner sep=3pt, minimum height=7mm, align=center},
  qcirc/.style={draw, circle, thick, inner sep=1.4pt}
}

\def\BibTeX{{\rm B\kern-.05em{\sc i\kern-.025em b}\kern-.08em
    T\kern-.1667em\lower.7ex\hbox{E}\kern-.125emX}}
\begin{document}

\title{Quantum-Optimized Selective State Space Model for Efficient Time Series Prediction\\

}

\author{\IEEEauthorblockN{\c Stefan-Alexandru Jura, Mihai Udrescu, Alexandru Top\^{i}rceanu}
\IEEEauthorblockA{\textit{Department of Computer and Information Technology} \\
\textit{Politehnica University Timi\c soara}\\
Timi\c soara, Romania\\
\text{stefan.jura@student.upt.ro, mihai.udrescu@cs.upt.ro, alexandru.topirceanu@cs.upt.ro}}

}

\maketitle

\begin{abstract}
Long-range time series forecasting remains challenging, as it requires capturing non-stationary and multi-scale temporal dependencies while maintaining noise robustness, efficiency, and stability. Transformer-based architectures such as Autoformer and Informer improve generalization but suffer from quadratic complexity and degraded performance on very long time horizons. State space models, notably S-Mamba, provide linear-time updates but often face unstable training dynamics, sensitivity to initialization, and limited robustness for multivariate forecasting. To address such challenges, we propose the Quantum-Optimized Selective State Space Model (Q-SSM), a hybrid quantum-optimized approach that integrates state space dynamics with a variational quantum gate. Instead of relying on expensive attention mechanisms, Q-SSM employs a simple parametrized quantum circuit (RY–RX ansatz) whose expectation values regulate memory updates adaptively. This quantum gating mechanism improves convergence stability, enhances the modeling of long-term dependencies, and provides a lightweight alternative to attention. We empirically validate Q-SSM on three widely used benchmarks, i.e., ETT, Traffic, and Exchange Rate. Results show that Q-SSM consistently improves over strong baselines (LSTM, TCN, Reformer), Transformer-based models, and S-Mamba. These findings demonstrate that variational quantum gating can address current limitations in long-range forecasting, leading to accurate and robust multivariate predictions.
\end{abstract}


\section{Introduction}
Time series forecasting is a statistical and machine learning approach that uses past data to predict future values of a series---it stems from the classical philosophical problem of induction \cite{montgomery2015introduction}. Time series analysis finds applications in a wide range of domains, from energy systems to finance, climate, transportation, healthcare, politics, and more. In all fields, forecast quality is critical for effective decision-making. For example, accurate predictions of electricity demand help run a stable grid and ensure optimal use of renewable sources; exchange rate forecasting assesses risks and plans economic activities; weather and environment forecasts secure population safety with timely warning. However, in practice, all these series are noisy and non-stationary, and they are also often multivariate, meaning that several time signals are necessary to predict the future signal \cite{box2015time}. Such issues make forecasting challenging when both short-term variations and long-range signals must be taken into account, while the computational process should be efficient. Indeed, computational efficiency is fundamental in forecasting, as time steps of interest can extend to days, weeks, or even months in some applications.

Conventional time series forecasting methods, including statistical models (e.g., ARIMA, VAR), and recently, early deep learning based methods (e.g., LSTM, GRU, TCN) have been proven to be effective for short-term forecasting \cite{box1970arima,hochreiter1997lstm,cho2014gru,bai2018tcn}. Nevertheless, due to limited memory and predominant sequential computation, it is challenging to learn long-range temporal dependencies with these models; they do not work well on long prediction horizons, which makes them less competitive on real-world problems where long-term forecasting accuracy is required. On the other hand, these model forms may lack generality in capturing some more general $n$-dimensional array of series with complicated cross-term interactions \cite{box2015time, bai2018tcn}. Thus, there is a demand for forecasting models that can model short-term dynamics while achieving a consistently stable performance prediction for long-term horizons at scale.

The latest developments in long-term forecasting focus on two main directions: Transformer-based architectures \cite{vaswani2017transformer,wu2021autoformer,zhou2021informer,zhou2022fedformer} and state space models (SSMs) \cite{gu2022s4,wang2024simplemamba}. Transformer-based variants, such as Informer, Autoformer, and Reformer, introduced strategies dealing with long sequences, though their dependencies on the operations of attention generate quadratic time and memory burden \cite{wu2021autoformer,zhou2021informer,kitaev2020reformer}; such an overhead renders them not practically useful for extremely long horizons and very large datasets. Moreover, there is a decrease in their predictive performance in the wake of increases in the forecasting horizon. Alternatively, SSM-based models—S-Mamba being the most recent—provide an attractive alternative, owing to their linear time inference via selectively updating the state. However, such models can suffer from unstable training dynamics, poor hyper-parameter initialization, and increased model sensitivity to initialization, as well as their inability to produce robust results when applied to multivariate time series data \cite{gu2023mamba,wang2024simplemamba}. Accordingly, neither the Transformers nor SSMs provide a complete resolution of the trade-off between accuracy, stability, and efficiency in long-range forecasting. To address these limitations, we introduce the Quantum-Optimized Selective State Space Model (Q-SSM), a hybrid quantum-classical model that leverages a selective state space dynamics extension, with a variational quantum circuit (VQC) gating mechanism. Q-SSM is superior to Transformers as it does not call for quadratic attention operations, and to S-Mamba because it stabilizes optimization with quantum adaptive memory control. Our Q-SSM architecture leads to higher convergence, robustness, and prediction accuracy (for both long and short-term horizons).

The main contributions of this paper are:
\begin{itemize}
    \item We present Q-SSM, the first q-optimized SSM with a VQC-based gating to control memory update on long sequences dynamically;
    \item We design a hybrid quantum-classical architecture where temporal dynamics are modeled through a selective state update regulated by a quantum gate, followed by a lightweight feed-forward decoder with residual forecasting; this design avoids heavy attention mechanisms, remains computationally efficient, and is well-suited for long-horizon multivariate prediction;
    \item We carry out extensive experiments on three commonly applied benchmarks—ETT, Electricity Load Diagrams, and Exchange Rate, which belong to different domains and contain short-term as well as long-term forecasting horizons;
    \item We show that Q-SSM consistently overperforms Transformer-based baselines (Autoformer, Informer, Reformer) and another state space baseline (S-Mamba), obtaining new state-of-the-art results in MAE and MSE.
\end{itemize}

\section{Background}
\subsection{Classical Statistical Models}

Traditional statistical approaches have been extensively studied for time series forecasting. The most prominent example is the \textbf{ARIMA} model~\cite{box2015time}, which combines autoregression, differencing, and moving average components. ARIMA works well for short-term and stationary series, but is unable to capture nonlinear patterns or long-range dependencies.  

The \textbf{Vector Autoregression (VAR)} model~\cite{sims1980var} generalizes autoregression to multivariate time series by expressing each variable as a linear function of its own past values and the lags of other variables. While useful in econometrics, VAR rapidly becomes computationally expensive and unstable as the number of variables increases.  

Other methods, such as \textbf{Exponential Smoothing (ETS)}~\cite{hyndman2008forecasting}, directly model level, trend, and seasonality components. ETS is simple and effective for univariate seasonal data, but it cannot capture nonlinear cross-variable dynamics.    

Overall, statistical models are computationally efficient and interpretable, yet their \textit{linear setting} limits their effectiveness for complex, nonlinear, and long-horizon forecasting tasks.

\subsection{Early Deep Learning Models}
The scientific community adopted neural networks for forecasting because they capture nonlinear relations and learn representations beyond classical statistical models. Recurrent neural networks (\textbf{RNNs}) were the early default choice \cite{rumelhart1986rnn}, but suffer from vanishing/exploding gradients on long contexts; \textbf{LSTM} \cite{hochreiter1997lstm} and \textbf{GRU} \cite{cho2014gru} mitigate this with gating and have been widely used in energy, finance, and traffic forecasting \cite{lim2021timeseries}. \textbf{TCNs} replace recurrence with 1-D dilated causal convolutions, enlarging the receptive field while enabling parallel training and stable gradients \cite{bai2018tcn}. Despite these gains, RNNs remain sequential (slow for very long histories) and TCNs require deeper stacks or larger dilations; both can be sensitive to hyperparameters and may underperform as the horizon grows, especially in high-dimensional multivariate settings. These limits motivated the move toward architectures that preserve long-range dependencies with better scalability, notably Transformer variants and modern State Space Models.

\subsection{Novel models---Transformers and State Space Models}

The Transformer architecture \cite{vaswani2017transformer} proposed a \textit{self-attention} mechanism that captures pairwise interactions between all tokens in a sequence. The attention mechanism forms an $L \times L$ similarity matrix for a sequence of length $L$ and allows the model to learn global interactions without losing information. This property allowed Transformers to outperform in NLP and (subsequently) forecasting, but also introduced a quadratic cost in both time and memory, which is problematic for really long time series.

To mitigate such scalability issues, several Transformer variants have been proposed. \textbf{Informer}\cite{zhou2021informer} mitigates attention cost with \emph{ProbSparse attention} that only calculates the attention between queries with high informativeness instead of a full attention map, combined with a generative-style decoder. \textbf{Autoformer}\cite{wu2021autoformer} introduces an \textit{auto-correlation mechanism} instead of attention, which captures periodic patterns among time series and decomposes input into trend and seasonal parts, especially fitting for long-term prediction. The \textbf{Reformer}~\cite{kitaev2020reformer} uses \textit{locality-sensitive hashes (LSH)} to approximate attention, reducing the complexity from quadratic to almost linear and reversible residual layers to minimize memory usage during training. Despite these advancements, Transformer-based models exhibit inaccuracy on very long horizons and costly space requirements for multivariate forecasting.

\textbf{State Space Models (SSMs)} model continuous-time dynamics via linear ordinary differential equations of the form
\begin{equation}
\label{eq:ssm}
    \frac{d}{dt} h(t) = A h(t) + B x(t), \quad y(t) = C h(t) + D x(t),
\end{equation}
where $A, B, C,$ and $D$ are the state-space matrices; $A$ is the state transition matrix, $B$ the input-to-state matrix, $C$ the observation matrix, and $D$ the feed-through matrix. $x(t)$ is the input vector at time $t$, $y(t)$ the output vector at time $t$, $h(t)$ the hidden state vector at time $t$, which summarizes past information. For practical time series forecasting, these dynamics are discretized, leading to linear recurrences of the form $h_{t+1} = \bar{A} h_t + \bar{B} x_t$, $y_t = \bar{C} h_t + \bar{D} x_t$. The discretization enables \textit{linear-time inference}, since updates are proportional to the sequence length. Deep variants such as \textbf{S4~\cite{gu2021s4}} introduced structured state matrices parameterized by HiPPO (High-order Polynomial Projection Operators) that allow for stable and efficient modeling of long dependencies.

\textbf{Mamba}~\cite{gu2023mamba} generalizes SSMs by allowing selective state updates. Rather than following pre-defined fixed transition matrices as in S4, Mamba learns to modulate the amount of hidden state updating and preserving at each time step. Such mechanisms are introduced through gating functions and nonlinear operations (e.g., sigmoid and softplus) to ensure the decay and the input influence, respectively, in a stable way; this allows the model to attend to the most important parts of the sequence while ignoring less useful signals. In contrast to classical SSMs, Mamba offers stronger long-range context modeling and is more efficient by performing linear-time inference, outperforming Transformer-based approaches on the long sequence benchmarks.

S-Mamba~\cite{wang2024simplemamba} was recently proposed as a specialized adaptation of Mamba for time-series forecasting. Its architecture integrates four main components designed to handle multivariate and long-horizon prediction tasks efficiently. First, each variable in the input sequence is linearly tokenized into an embedding space, which ensures feature normalization and prepares the data for sequential modeling. Then, a bidirectional selective state space block captures variable correlations by processing information both forward and backward, allowing inter-variate dependency modeling without explicit attention mechanisms. On top of this, a feed-forward network (FFN) with residual connections and normalization enhances non-linearity and strengthens the representation of temporal dependencies. Finally, a linear projection layer maps the learned representations to the prediction horizon, enabling multi-step forecasting across different settings.

Compared to Transformer-based approaches, S-Mamba provides more efficient memory usage and avoids quadratic complexity while achieving competitive state-of-the-art results on long-term forecasting benchmarks. It also improves stability over the original Mamba model, particularly in multivariate contexts. However, S-Mamba still has limitations: it remains sensitive to hyperparameter initialization, requires careful tuning, and introduces extra training costs due to its bidirectional block and FFN layers. As a result, while it narrows the gap between scalability, stability, and accuracy, S-Mamba does not fully resolve this trade-off for long-range time series forecasting.

\subsection{Quantum Computing}
Quantum computing (QC) is a computational paradigm that abides by the laws of quantum mechanics rather than classical physics. In QC, the fundamental unit of information is the quantum bit or \emph{qubit}. A qubit is a unit vector in a two–dimensional complex Hilbert space \cite{prodan2007design, nielsen2010quantum},
\begin{equation}\label{eq:qubit-state}
|\psi\rangle=\alpha|0\rangle+\beta|1\rangle,\qquad |\alpha|^2+|\beta|^2=1.
\end{equation}
The qubit measurement on the computational basis $\left\{\vert 0\rangle , \vert 1\rangle\right\}$ renders $\vert 0\rangle$ or $\vert 1\rangle$ with probabilities $\vert\alpha\vert^2$ and $\vert\beta\vert^2$, respectively. An $N$-qubit state can be in a superposition of all its possible values, 
\begin{equation}
    \label{eq:qstate}
    \vert\psi\rangle = \sum_{i=0}^{2^N-1} a_i\vert i\rangle, 
\end{equation}
where $a_i$ are the quantum amplitudes, with $\sum_{i=0}^{2^N-1}\vert a_i\vert = 1$.

The quantum computational process consists of changing quantum states using unitary transformations $U$. Simpler unitary transformations are called quantum gates; quantum circuits assemble such quantum gates to implement more complex unitary transformations. For single-qubit gates, parameterized rotations around Pauli axes form a convenient gate set; in this paper, we use
\begin{equation}\label{eq:rotations}
|\psi(\theta,\phi)\rangle = R_X(\phi)\,R_Y(\theta)\,|0\rangle,
\end{equation}
with $R_{X/Y}(\cdot)=\exp\!\left(-\tfrac{i}{2}(\cdot)\,X/Y\right)$. In quantum computing, an \emph{observable} is a mathematical operator (specifically a Hermitian matrix) that represents a physical property of the system which can be measured. For example, the Pauli matrices $X, Y, Z$ are standard observables corresponding to spin projections along the three axes. When we measure an observable on a quantum state, the outcome is one of its eigenvalues, while the expectation value gives the average result over many measurements \cite{preskill1998lecture}.
Expectation values of observables provide smooth, bounded nonlinearities. In particular, the Pauli-$Z$ expectation of the rotated state is
\begin{equation}\label{eq:zexp}
\langle Z\rangle
=\langle \psi(\theta,\phi)|Z|\psi(\theta,\phi)\rangle
=\cos\theta\,\cos\phi\in[-1,1].
\end{equation}

Gradients of such expectations are obtained analytically via the parameter-shift rule, enabling end-to-end training on state-vector simulators:
\begin{equation}\label{eq:param-shift}
\frac{\partial}{\partial \theta}\langle Z\rangle
=\tfrac{1}{2}\Big(\langle Z\rangle_{\theta+\frac{\pi}{2}}-\langle Z\rangle_{\theta-\frac{\pi}{2}}\Big).
\end{equation}
Our experiments rely solely on single-qubit rotations, projective measurement, and differentiable expectations, so that no multi-qubit circuitry is required \cite{nielsen2010quantum,schuld2019quantum}.

\subsection{Quantum Machine Learning}
\textbf{Quantum Machine Learning (QML)} is a field that combines conventional learning machines with the power of quantum computing. QML is mainly applied to domains such as chemistry, finance, and combinatorial optimization~\cite{biamonte2017quantum,schuld2019quantum}; its application in Time Series Prediction (TSP) has been explored less. This provides a strong motivation for hybrid quantum-classical models that exploit quantum gating to enhance stability and generalization in sequence learning. The rationale is that parametrized quantum circuits can represent complex, high-dimensional functions with favorable optimization properties. 

A core building block in Quantum Machine Learning (QML) is the \textbf{parametrized quantum circuit} (PQC), sometimes referred to as a variational quantum circuit. In such models, a quantum state is prepared by applying a sequence of parametrized unitary rotations, e.g.,
\begin{equation}
    |\psi(\boldsymbol{\theta})\rangle = U(\boldsymbol{\theta})|0\rangle,
\end{equation}
where $\boldsymbol{\theta}$ denotes a vector of tunable parameters. Observables are then measured, typically yielding expectation values such as
\begin{equation}
    f(\boldsymbol{\theta}) = \langle \psi(\boldsymbol{\theta}) | O | \psi(\boldsymbol{\theta}) \rangle,
\end{equation}
which represents the expectation value of the observable $O$. Unlike the discrete outcomes of projective measurements, this expectation value is a smooth and differentiable function of the circuit parameters $\boldsymbol{\theta}$, making it directly usable within gradient-based learning algorithms.

The parameters $\boldsymbol{\theta}$ are trained by minimizing a classical cost function, often related to prediction error or energy expectation. Crucially, modern QML frameworks (e.g., PennyLane, Qiskit, TensorFlow Quantum) support automatic differentiation of PQCs; this is enabled through the \textit{parameter-shift rule}, which provides unbiased gradient estimates for expectation values of quantum circuits. As a result, PQCs can be optimized end-to-end with conventional gradient-based optimizers such as Adam or SGD, in the same way as classical neural networks~\cite{schuld2020circuit,cerezo2021variational}.

The variational principle has been extensively studied as \textbf{variational quantum algorithms} (VQAs)~\cite{cerezo2021variational}, with applications in quantum chemistry, combinatorial optimization, and recently, machine learning. In ML, PQCs can act as nonlinear feature maps~\cite{schuld2019quantum} or gating mechanisms, due to their smooth and bounded functional form, which naturally avoids gradient explosion or vanishing.

\section{Proposed architecture}
\subsection{Overview}
Our proposed model, \textbf{Quantum-Optimized Selective State Space Model (Q-SSM)}, addresses the inherent trade-off between scalability, stability, and accuracy in long-range time series forecasting. Figures \ref{fig:encoder} and \ref{fig:decoder} present our model's architecture, consisting of three major components: 

\begin{itemize}
    \item \textbf{Input Encoder}, responsible for preparing the raw multivariate time series by incorporating the original measurements and additional calendar features (e.g., hour-of-day, day-of-year, and their sinusoidal encodings); this ensures that seasonal and periodic structures are explicitly represented in the input space.
    
    \item \textbf{Selective State Space Backbone with Quantum Gate}, a linear-time recurrent mechanism that processes the input sequence. Unlike classical state space models that rely on fixed or purely data-driven gating functions, our backbone integrates a \emph{variational quantum gate}. This gate adaptively regulates memory updates, ensuring that the hidden state selectively retains or discards past information stably. The quantum gate delivers smooth, bounded, and differentiable outputs, stabilizing training and mitigating gradient vanishing or explosion.
    
    \item \textbf{Forecasting Decoder}, a lightweight feed-forward projection that maps the final hidden state into the prediction horizon. To further stabilize predictions and improve accuracy, the decoder employs a residual formulation, where the last observed values are added back to the forecast.
\end{itemize}

Overall, the Q-SSM architecture avoids the quadratic complexity of Transformer-based models by adopting a linear recurrence, while overcoming the instability issues of purely classical state space models through quantum gating. This results in an efficient, robust, and accurate forecaster for multivariate and long-horizon time series.

The input to our model is a sliding window of multivariate observations: $X \in \mathbb{R}^{B \times T \times F}$, where $B$ is the batch size, $T$ is the input sequence length, and $F$ is the number of features. Each instance corresponds to a fixed-length historical segment used to predict the subsequent horizon.

In addition to the original data provided by the datasets (such as temperature, load, or exchange rates), we augment the features in our model explicitly with \textbf {calendar features} that capture regular temporal structure. For example, for the ETT datasets, we append the following characteristics: \begin{itemize} \item \emph Hour of the day: $\sin (2\pi \cdot \hbox {hour}/24)$ and $\cos(2\pi \cdot \hbox {hour}/24)$, capturing the cyclical nature of daily patterns. \item \emph Day of the year: $\sin (2\pi \cdot \hbox {day}/365)$ and $\cos(2\pi \cdot \hbox {day}/365)$, capturing seasonal variation over the annual cycle. \end{itemize}

Through this encoding, the model has access to the raw multivariate signals and an explicit representation of the periodicity of time; this combination allows for the capture of both short-term fluctuations and long-range seasonal dependencies, which are decisive factors for practical forecasting problems.

\subsection{State Space Backbone}
The core idea of Q-SSM is a \textbf{selective state space backbone} that models sequential dependencies in linear time. Let $x_t \in \mathbb{R}^F$ denote the feature vector at time step $t$. The backbone maintains a hidden state $h_t \in \mathbb{R}^d$, which memorizes past information and is updated according to the gated recurrence
\begin{equation}
\label{eq:recgat}
h_t = (1 - g)\,h_{t-1} + g \cdot \text{LN}\big(W(P(x_t)) + b + \alpha \cdot c\big),
\end{equation}
where:
\begin{itemize}
    \item $P: \mathbb{R}^F \to \mathbb{R}^k$ is a linear projection that maps the input features to an intermediate embedding space of dimension $k$.
    \item $W: \mathbb{R}^k \to \mathbb{R}^d$ further transforms the embedding into the hidden dimension $d$.
    \item $b \in \mathbb{R}^d$ is a learnable bias term.
    \item $\alpha \in \mathbb{R}$ is a learnable scalar that controls the influence of the aggregated \textbf{calendar signal} $c$ (mean value of the calendar features across the input window).
    \item LN denotes Layer Normalization, which stabilizes optimization and improves convergence.
    \item $g \in [0.05, 0.95]$ is the gating coefficient produced by the quantum module (described in the next subsection).
\end{itemize}

The update in Equation \ref{eq:recgat} ensures that the hidden state $h_t$ adaptively balances between retaining information from the previous step $(1-g)\,h_{t-1}$ and integrating new input-driven information $g \cdot \text{LN}(\cdot)$. The inclusion of the calendar scalar provides an additional global temporal context, while Layer Normalization prevents internal covariate shifts.

Unlike classical RNNs or LSTMs, which suffer from vanishing or exploding gradients due to repeated multiplications, the proposed backbone operates with additive gated updates. This structure entails \textbf{linear-time complexity} with respect to sequence length $O(T)$, making the model scalable to very long input horizons. Furthermore, the selective update mechanism provides the foundation upon which the quantum gate can exert fine-grained adaptive control over memory dynamics.

\subsection{Quantum Gate}
A central component of Q-SSM is its gating mechanism, which regulates the memory update,
\begin{equation}\label{eq:qssm-update}
h_t = (1-g)\,h_{t-1} + g\,u_t.
\end{equation}
In Equation \ref{eq:qssm-update}, $g\in (0,1)$ to interpolate between retaining the past $h_{t-1}$ and incorporating the new input $u_t$. Classical recurrent models define
\begin{equation}\label{eq:classical-gate}
g_{\text{classical}} = \sigma(w^\top x_t + b),
\mbox{ where }
\sigma(x) = \frac{1}{1+e^{-x}} .
\end{equation}
Although this guarantees $g\in(0,1)$, it has two limitations for long-horizon forecasting:
\begin{itemize}
    \item[(i)] \textbf{Vanishing gradients.} The derivative $\sigma'(x) = \sigma(x)(1-\sigma(x))$ decays exponentially as $|x|$ grows, i.e., $\sigma'(x) \leq \tfrac{1}{4}$ and $\sigma'(x)\to 0$ for $|x|\gg 0$. This prevents adaptive updates once the gate saturates.  
    \item[(ii)] \textbf{Linear pre-activation.} The argument $w^\top x_t+b$ is a linear map of the input, limiting the richness of gating dynamics and causing poor adaptability in non-stationary settings.  
\end{itemize}

\subsubsection{Quantum-enhanced pre-activation}

We propose to replace the linear pre-activation with the expectation of parametrized quantum circuits. A single qubit is initialized to $\vert0\rangle$, rotated by
\begin{equation}\label{eq:state-prep}
|\psi(\theta,\phi)\rangle = R_X(\phi)\,R_Y(\theta)\,|0\rangle,
\end{equation}
and then measured on the $Z$ basis, yielding
\begin{equation}\label{eq:expval-z}
z(\theta,\phi) = \langle \psi(\theta,\phi)|Z|\psi(\theta,\phi)\rangle 
= \cos\theta \cos\phi.
\end{equation}
This nonlinear, oscillatory function takes values in $[-1,1]$. Two independent circuits produce the linearly combined $z_1$ and $z_2$,
\begin{equation}\label{eq:quantum-logit}
s = w_1 z_1 + w_2 z_2 + b_g .
\end{equation}
The final gate is
\begin{equation}\label{eq:gate-clipped}
g = \mathrm{clip}\!\left(\sigma(s),\,g_{\min},\,g_{\max}\right),
\end{equation}
with $0< g_{\min}<g_{\max}<1$ (empirically, $g_{\min}=0.05, g_{\max}=0.95$). The clipping guarantees that the gate value $g$ cannot reach the degenerate extremes $0$ (full memory retention, no update) or $1$ (full overwrite, no memory), which would otherwise harm stability.

\subsubsection{Bounded derivatives and Lipschitz continuity}
The derivatives of the quantum expectation are
\begin{equation}\label{eq:dz-partials}
\frac{\partial z}{\partial \theta} = -\sin\theta \cos\phi,\qquad 
\frac{\partial z}{\partial \phi} = -\cos\theta \sin\phi.
\end{equation}
As such, we have
\begin{equation}\label{eq:dz-bounds}
\Big|\frac{\partial z}{\partial \theta}\Big| \leq 1,\quad 
\Big|\frac{\partial z}{\partial \phi}\Big| \leq 1.
\end{equation}
This implies that $z(\theta,\phi)$ is 1-Lipschitz w.r.t. each parameter. For the quantum-parameterized gate $g$ in Eq.~\eqref{eq:gate-clipped}, its gradients with respect to the circuit parameters follow from the chain rule,
\begin{equation}\label{eq:dg-chain}
\frac{\partial g}{\partial \theta_i} = \sigma'(s)\,w_i\,\frac{\partial z_i}{\partial \theta_i}, \qquad
\frac{\partial g}{\partial \phi_i} = \sigma'(s)\,w_i\,\frac{\partial z_i}{\partial \phi_i}.
\end{equation}
Since $\sigma'(s) \leq 1/4$ and $\bigl|\dfrac{\partial z}{\partial \theta_i}\bigr|,\,\bigl|\dfrac{\partial z}{\partial \phi_i}\bigr|\leq 1$, we obtain the Lipschitz bound
\begin{equation}\label{eq:dg-bounds}
\Big|\frac{\partial g}{\partial \theta_i}\Big| \leq \dfrac{|w_i|}{4}, \qquad
\Big|\frac{\partial g}{\partial \phi_i}\Big| \leq \dfrac{|w_i|}{4}.
\end{equation}
Therefore, $g$ is globally Lipschitz continuous in the quantum parameters, which ensures stable gradients during optimization, preventing both explosion and collapse.

\subsubsection{Contractivity of memory update}
The recurrence Jacobian w.r.t. $h_{t-1}$ is
\begin{equation}\label{eq:jacobian}
\frac{\partial h_t}{\partial h_{t-1}} = (1-g)I,
\end{equation}
where $I$ is the identity matrix. Because $g\in[g_{\min},g_{\max}]$ with $0<g_{\min}<g_{\max}<1$, we guarantee
\begin{equation}\label{eq:jacobian-norm}
\biggl\| \dfrac{\partial h_t}{\partial h_{t-1}} \biggr\|_2 = |1-g| < 1.
\end{equation}
Hence, the update is a contraction mapping. This property guarantees that information is neither explosively amplified nor completely forgotten, enabling stable memory over long sequences.

\subsubsection{Why quantum and sigmoid rather than sigmoid alone}
It may seem redundant to apply $\sigma(\cdot)$ after a quantum function $z(\theta,\phi)$. However, the crucial difference lies in the \emph{geometry of the pre-activation}:
\begin{itemize}
    \item In classical gates, $s = w^\top x+b$ is linear in $x$, so $\sigma(s)$ has limited flexibility;  
    \item In quantum gates, $s = w_1 \cos\theta_1\cos\phi_1 + w_2 \cos\theta_2\cos\phi_2 + b_g$ is oscillatory and nonlinear in trainable parameters; this creates a richer family of input distributions for the sigmoid, avoiding trivial saturation.  
\end{itemize}
Thus, the sigmoid acts not as the source of nonlinearity but as a \emph{normalization step} mapping the expressive quantum features into $(0,1)$. The clipping further avoids degeneracy.

\subsubsection{Impact on time series}

Time series combine \emph{slow components} (trends, seasonality) with \emph{fast variations} (shocks, noise). The quantum gate adapts between these regimes,
\begin{itemize}
    \item Small $g \approx g_{\min}$: the update is slow, $h_t \approx h_{t-1}$, preserving long-term memory of seasonal patterns;  
    \item Large $g \approx g_{\max}$: the update is fast, $h_t \approx u_t$, allowing rapid adaptation to short-term fluctuations.  
\end{itemize}
Because $g$ is always kept strictly between 0 and 1, the model continuously blends old and new information, instead of collapsing into either rigid memory retention or complete forgetting. This balance is especially advantageous in multivariate, non-stationary, and long-horizon forecasting. In practice, the gate only requires the simulation of two single-qubit expectation values per batch, which is computationally negligible compared to the linear layers. This choice ensures that the quantum component remains lightweight and stable, while still injecting non-classical smoothness into the gating dynamics. Significantly, the parameters $\theta, \phi, w_1, w_2, b_g$ are all updated through backpropagation using the parameter-shift rule from Equation \ref{eq:param-shift}, thus enabling end-to-end training together with the rest of the model.

\begin{figure}[h]
    \centering
    \includegraphics[scale=0.7]{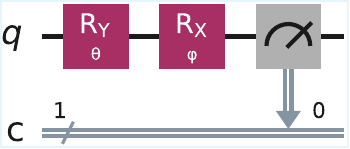}
    \caption{Quantum gating circuit: the qubit, initialized in $|0\rangle$, 
    undergoes successive rotations $R_Y(\theta)$ and $R_X(\phi)$ before being measured 
    in the $Z$-basis. The classical wire \texttt{c} shown below the quantum wire is 
    automatically added by Qiskit to indicate the measurement outcome stored in a 
    one-bit classical register. In practice, we do not use the discrete 0/1 result, 
    but rather the expectation value $\langle Z \rangle \in [-1,1]$, which is then 
    passed through a sigmoid function to form the adaptive gating coefficient $g$.}
    \label{fig:quantum_gate}
\end{figure}

\subsection{Decoder and Forecast Reconstruction}

Let $h \in \mathbb{R}^d$ denote the latent state vector obtained from the quantum-gated backbone. The goal of the decoder is to map this compact representation into a multivariate forecast,
\begin{equation}
\label{eq:yhat}
\hat{Y} \in \mathbb{R}^{H \times F},
\end{equation}
where $H$ is the prediction horizon (number of future time steps) and $F$ is the number of variables (features) in the input sequence.

The decoder operates in four steps:

\begin{enumerate}
    \item \textbf{Nonlinear projection.}  
    The latent state $h$ is first projected into a higher-level representation $z \in \mathbb{R}^d$ via a Multi-Layer Perceptron (MLP),
    \[
        z = \mathrm{ReLU}(W_1 h + b_1),
    \]
    where $W_1 \in \mathbb{R}^{d \times d}$ and $b_1 \in \mathbb{R}^d$ are learnable parameters and $\mathrm{ReLU}(x)=\mathrm{max}(0,x)$.
    This operation introduces non-linearity and enriches the latent representation.

    \item \textbf{Regularization.}  
    To prevent overfitting, we apply a dropout operator with drop probability $p=0.1$,
    \[
        z' = \mathrm{Dropout}(z; p=0.1).
    \]
    This randomly sets some entries of $z$ to zero during training while leaving inference unchanged.

    \item \textbf{Forecast projection.}  
    The regularized vector $z'$ is linearly mapped to the flat forecast vector
    \[
        \hat{y}_{\text{flat}} = W_2 z' + b_2,
    \]
    where $W_2 \in \mathbb{R}^{(H \cdot F) \times d}$ and $b_2 \in \mathbb{R}^{H \cdot F}$ are learnable parameters.  
    The result is reshaped into the two-dimensional matrix
    \[
        \hat{Y}_{\text{base}} \in \mathbb{R}^{H \times F},
    \]
    where each row corresponds to a future time step and each column corresponds to a variable.

    \item \textbf{Residual correction.}  
    Let $x_T \in \mathbb{R}^F$ denote the last observed input vector (the most recent time step of the encoder sequence).  
    The final forecast is obtained as
    \[
        \hat{Y} = \hat{Y}_{\text{base}} + \mathbf{1}_H x_T^\top,
    \]
    where $\mathbf{1}_H \in \mathbb{R}^{H \times 1}$ is a vector of ones.  
    This residual formulation ensures that predictions are centered around the last observation, and the decoder only needs to model deviations from it.
\end{enumerate}

\noindent

Predicting absolute values often causes long-term drift, especially for non-stationary time series. By adding forecasts to $x_T$ and predicting only relative changes, our model:
\begin{itemize}
    \item reduces error accumulation across long horizons;  
    \item improves stability by grounding predictions in the most recent observation;  
    \item makes learning easier since the decoder works on modeling differences rather than absolute magnitudes.
\end{itemize}

\subsection{Complexity Analysis}

The computational cost of Q--SSM can be separated into the recurrent backbone and the decoder components.  
At each step $t \in \{1,\dots,W\}$, the hidden state is updated as $h_t = (1-g)\,h_{t-1} + g \cdot f(x_t)$, $f(x_t)=\mathrm{LN}(W_2(W_1x_t))$,
where $W_1\in\mathbb{R}^{k\times F}$ and $W_2\in\mathbb{R}^{d\times k}$.  
This projection costs $\mathcal{O}(Fk+kd)$, leading to $\mathcal{O}(W(Fk+kd))$ over a sequence of length $W$. The recurrence is therefore linear in $W$ and bilinear in the feature ($F$), hidden ($d$), and projection ($k$) dimensions.  

The decoder takes the final hidden state $h\in\mathbb{R}^d$ and generates $H\times F$ predictions via a two-layer MLP, which requires $\mathcal{O}(HdF)$ operations. The overall complexity is thus $\mathcal{O}(W(Fk+kd) + HdF)$.

In practice, the number of input features $F$ is dataset-dependent, see Section \ref{sec:ex-set}. This variation changes the constants in runtime but not the asymptotic order. For ETT and Traffic, seasonal encodings ($\sin/\cos$ of hour, day, week) add a small constant to $F$, while Exchange remains dominated by non-periodic stochastic features.

Compared to baselines, Q--SSM avoids the $\mathcal{O}(W\log W)$ scaling of Autoformer~\cite{wu2021autoformer}, which grows superlinearly with sequence length, and reduces the quadratic $\mathcal{O}(Wd^2)$ dependence of Mamba~\cite{gu2023mamba} on hidden dimension. The quantum gate adds only a constant overhead (one-qubit expectations per step), negligible under simulation relative to the matrix multiplications in the backbone and decoder.

Overall, Q--SSM maintains linear scaling in both window length $W$ and forecast horizon $H$, adapts efficiently to datasets of very different feature dimensionality, and provides a favorable trade-off between expressiveness and computational cost.

\begin{figure}[t]
\centering
\begin{tikzpicture}[
  node distance=11mm,
  every node/.style={font=\sffamily\footnotesize},
  block/.style={draw, rounded corners, thick, minimum width=30mm, minimum height=9mm, align=center},
  bigop/.style={draw, circle, thick, minimum size=10mm, align=center, font=\Large},
  arr/.style={-Stealth, thick}
]

\node[block, fill=blue!15] (inp) {Input window \\ $X \in \mathbb{R}^{W \times F}$};
\node[block, fill=green!15, below=of inp] (time) {Time features \\ $\sin/\cos$};
\node[block, fill=orange!20, below=of time] (proj1) {Projection $P:\; \mathbb{R}^F \to \mathbb{R}^k$};
\node[block, fill=orange!30, below=of proj1] (proj2) {Projection $W:\; \mathbb{R}^k \to \mathbb{R}^d$};
\node[block, fill=purple!15, below=of proj2] (ln) {LayerNorm \\ $u_t \in \mathbb{R}^d$};

\node[bigop, below=of ln] (mult) {$\otimes$};
\node[bigop, below=of mult] (sum) {$\Sigma$};

\node[block, fill=gray!15, below=of sum] (ht) {Updated state \\ $h_t \in \mathbb{R}^d$};

\node[block, fill=cyan!20, right=14mm of mult] (qgate) {Quantum gate \\ $R_Y(\theta), R_X(\phi)$ \\ $g \in (0,1)$};

\draw[arr] (inp) -- node[right] {$X \in \mathbb{R}^{W \times F}$} (time);
\draw[arr] (time) -- node[right] {$X' \in \mathbb{R}^{W \times (F+\text{no. of calendar features})}$} (proj1);
\draw[arr] (proj1) -- node[right] {$v_t \in \mathbb{R}^k$} (proj2);
\draw[arr] (proj2) -- node[right] {$z_t \in \mathbb{R}^d$} (ln);
\draw[arr] (ln) -- node[right] {$u_t \in \mathbb{R}^d$} (mult);
\draw[arr] (mult) -- node[right] {$g \cdot u_t$} (sum);
\draw[arr] (sum) -- node[right] {$h_t \in \mathbb{R}^d$} (ht);

\draw[arr] (qgate.west) -- node[above] {$g$} (mult.east);

\node[left=18mm of sum] (htm1) {$h_{t-1} \in \mathbb{R}^d$};
\draw[arr] (htm1.east) -- (sum.west);

\end{tikzpicture}
\caption{Encoder block with quantum gating. The input $X$ is augmented with time features, projected into latent spaces $k$ and $d$, normalized to $u_t$, and modulated by a quantum-derived gate $g$. The product $g \otimes u_t$ is summed with the previous state $h_{t-1}$ to yield the updated hidden state $h_t$.}
\label{fig:encoder}
\end{figure}
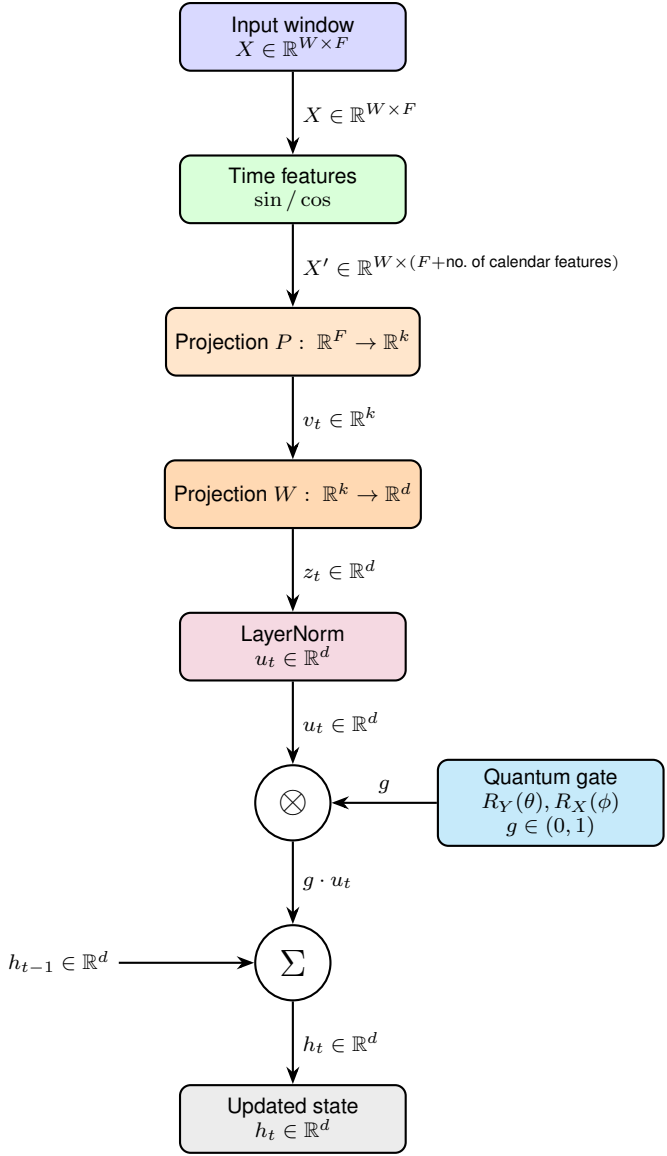
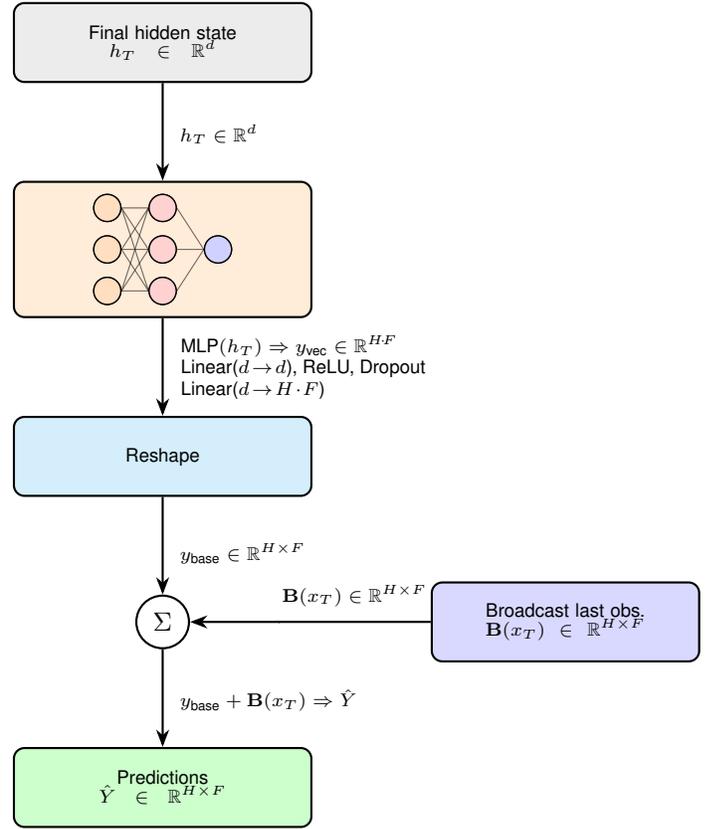
\begin{figure}[t]
\centering
\begin{tikzpicture}[
  scale=0.92, every node/.style={transform shape},
  node distance=13mm, >={Latex[length=2mm]},
  every node/.style={font=\sffamily\scriptsize},
  block/.style={draw, rounded corners, thick,
                minimum width=38mm, minimum height=10.5mm,
                align=center, text width=38mm, inner sep=2.2pt},
  bigop/.style={draw, circle, thick, minimum size=7mm, align=center, font=\normalsize},
  arr/.style={-Stealth, thick}
]

\node[block, fill=gray!15] (hT) {Final hidden state\\[-1pt]{\scriptsize $h_T \in \mathbb{R}^{d}$}};

\node[block, fill=orange!15, below=of hT, minimum height=18mm] (mlp) {};
\begin{scope}
  \pgfpathrectanglecorners{\pgfpointanchor{mlp}{south west}}%
                           {\pgfpointanchor{mlp}{north east}}
  \pgfusepath{clip}

  \coordinate (C) at (mlp.center);
  \foreach \y in {6,0,-6}{
    \node[circle, draw, fill=orange!25, minimum size=3.6mm] (l1\y) at ([xshift=-8mm,yshift=\y mm]C) {};
  }
  \foreach \y in {6,0,-6}{
    \node[circle, draw, fill=red!18, minimum size=3.6mm] (l2\y) at ([xshift=0mm,yshift=\y mm]C) {};
  }
  \node[circle, draw, fill=blue!18, minimum size=3.6mm] (lo) at ([xshift=8mm]C) {};
  \foreach \a in {6,0,-6}{\foreach \b in {6,0,-6}{\draw[line width=0.35pt, opacity=.35] (l1\a.east)--(l2\b.west);}}
  \foreach \b in {6,0,-6}{\draw[line width=0.35pt, opacity=.35] (l2\b.east)--(lo.west);}
  \foreach \y in {6,0,-6}{
    \node[circle, draw, fill=orange!25, minimum size=3.6mm] (l1\y) at ([xshift=-8mm,yshift=\y mm]C) {};
  }
  \foreach \y in {6,0,-6}{
    \node[circle, draw, fill=red!18, minimum size=3.6mm] (l2\y) at ([xshift=0mm,yshift=\y mm]C) {};
  }
  \node[circle, draw, fill=blue!18, minimum size=3.6mm] (lo) at ([xshift=8mm]C) {};
  \foreach \a in {6,0,-6}{\foreach \b in {6,0,-6}{\draw[line width=0.35pt, opacity=.35] (l1\a.east)--(l2\b.west);}}
  \foreach \b in {6,0,-6}{\draw[line width=0.35pt, opacity=.35] (l2\b.east)--(lo.west);}
\end{scope}

\node[block, fill=cyan!15, below=of mlp] (reshape) {Reshape};
\node[bigop, below=of reshape] (sum) {$\Sigma$};
\node[block, fill=green!20, below=of sum] (yhat) {Predictions\\[-1pt]{\scriptsize $\hat Y \in \mathbb{R}^{H\times F}$}};

\node[block, fill=blue!15, right=32mm of sum, text width=34mm, minimum width=34mm] (bx)
  {Broadcast last obs.\\[-1pt]{\scriptsize $\mathbf{B}(x_T)\in\mathbb{R}^{H\times F}$}};

\draw[arr] (hT) -- node[right, pos=0.52, xshift=1mm] {$h_T \in \mathbb{R}^{d}$} (mlp);
\draw[arr] (mlp) -- node[right, xshift=1.1mm, align=left]
  {$\text{MLP}(h_T)\Rightarrow y_{\text{vec}}\in\mathbb{R}^{H\!\cdot\!F}$\\
   Linear($d\!\to\!d$), ReLU, Dropout\\
   Linear($d\!\to\!H\!\cdot\!F$)} (reshape);
\draw[arr] (reshape) -- node[right, pos=0.58, xshift=1mm] {$y_{\text{base}}\in\mathbb{R}^{H\times F}$} (sum);
\draw[arr] (bx.west) -- ++(-22mm,0)
  node[midway, above=0.8mm] {$\mathbf{B}(x_T)\in\mathbb{R}^{H\times F}$}
  |- (sum.east);
\draw[arr] (sum) -- node[right, pos=0.52, xshift=1mm] {$y_{\text{base}}+\mathbf{B}(x_T)\Rightarrow \hat Y$} (yhat);

\end{tikzpicture}
\caption{Decoder with  MLP and residual connection. The MLP maps $h_T$ into $y_{\text{vec}}$, reshaped to $y_{\text{base}}$, then combined with the broadcast of the last observation to yield $\hat Y$.}
\label{fig:decoder}
\end{figure}

\section{Experimental Setup}
\label{sec:ex-set}
\subsection{Implementation details}

We implemented all models in Python using \texttt{PyTorch} for the classical backbone and \texttt{PennyLane} for the quantum gate. We conducted the experiments on Google Colab Pro with a single NVIDIA Tesla T4 GPU (16\,GB memory). The quantum module evaluates only two expectation values of a single simulated qubit, so its runtime overhead is negligible compared to the neural network forward and backward passes.

We trained using the Adam optimizer (initial learning rate $10^{-3}$, weight decay $10^{-4}$), with \texttt{ReduceLROnPlateau} (factor $0.5$, patience $3$) and early stopping (patience $10$). Unless otherwise specified, the loss is the mean squared error (MSE). All linear layers were initialized with the Kaiming (He) normal scheme and biases set to zero. This initialization preserves variance through ReLU layers, prevents vanishing/exploding activations, and stabilizes the residual decoder for long horizons. The gating parameters were initialized to start around $g \approx 0.5$ and are clamped in $[0.05,0.95]$, preventing degenerate cases where the update gate is always open or always closed. Random seeds were fixed for NumPy, PyTorch, and CUDA to ensure reproducibility.

Across datasets, we used the same backbone capacity: projection width $k=128$, latent width $d=128$, dropout $0.1$, lookback window $W=96$, and forecasting horizons $H \in \{96,192,336,720\}$. Forecasts are fully multivariate: the decoder outputs $H \times F$ predictions and adds back the last observed step via a residual connection. All reported metrics are computed on normalized data, consistent with prior long-horizon forecasting works \cite{wu2021autoformer,zhou2021informer,gu2023mamba,lim2023longterm}.

\subsection{Datasets and preprocessing}

We evaluate on four Electricity Transformer Temperature (ETT) variants \cite{haoyietal-informer-2021,zhou2021etdataset}, as well as the Traffic and Exchange Rate benchmarks \cite{lai2018lstnet,laiguokun_mts_data}. Each dataset is split chronologically into 60\% training, 20\% validation, and 20\% test. Standardization (z-score normalization) is fitted on the training set and applied to the validation and test. For all datasets, the model consumes all features and predicts all features (multivariate forecasting). Next, we explicitly describe the feature construction for each dataset:

\paragraph{ETT (ETTh1, ETTh2, ETTm1, ETTm2).}
The ETT datasets contain the target oil temperature (\texttt{OT}) and six exogenous load variables (\texttt{HUFL}, \texttt{HULL}, \texttt{MUFL}, \texttt{MULL}, \texttt{LUFL}, \texttt{LULL}), collected from Chinese electricity transformers between 2016 and 2018. To capture periodicity, we augment the raw features with four continuous calendar encodings: sine and cosine of hour-of-day, and sine and cosine of day-of-year. Thus, the input dimensionality is $F_{\text{ETT}} = 7 \text{ (raw)} + 4 \text{ (calendar)} = 11$.

The hourly subsets (ETTh1, ETTh2) contain 17,420 observations each, while the 15-minute subsets (ETTm1, ETTm2) contain 69,680 observations each. The window size is $W=96$ and horizons are $H \in \{96,192,336,720\}$.

\paragraph{Traffic.}
The Traffic dataset \cite{lai2018lstnet, laiguokun_mts_data} contains road occupancy rates (ranging from 0 to 1) measured every hour by 862 sensors deployed across the San Francisco Bay Area. Each time step, therefore, consists of $F=862$ correlated series, yielding a large-scale, high-dimensional forecasting problem. 

To capture strong daily and weekly periodicities, we augmented the raw sensor readings with calendar features: $\sin/\cos$ encodings of the hour of day and the day of week. These continuous embeddings preserve the cyclic nature of traffic patterns (e.g., rush hours, weekend vs. weekday dynamics) while avoiding discontinuous binary indicators. In contrast to ETT, where seasonal variation is primarily annual, Traffic exhibits short-term but highly regular periodic cycles driven by human activity. 

For consistency with prior long-horizon forecasting work, we use the same input length and horizon sizes as for the ETT benchmarks (i.e., window length $W=96$ and prediction horizons $H \in \{96, 192, 336, 720\}$). However, unlike ETT, where this corresponds to 24 hours or 15-minute intervals, in Traffic, these horizons represent 4 to 30 days ahead, which stresses the model’s ability to extrapolate periodic patterns across longer horizons in a highly multivariate setting.

\paragraph{Exchange Rate.}
The Exchange dataset consists of daily exchange rates of 8 currencies against the US Dollar from 1990 to 2016 \cite{laiguokun_mts_data}. Unlike ETT or Traffic, it exhibits weak seasonality and high stochasticity. Preliminary experiments confirmed that calendar encodings (e.g., day-of-week or month-of-year) did not improve performance. We therefore use the raw multivariate series without additional features: $F_{\text{Exchange}} = 8$. Here, we used the same window ($W=96$) and horizons ($H \in \{96,192,336,720\}$).

\subsection{Evaluation metrics}

We report the mean squared error (MSE) and mean absolute error (MAE) on normalized data. For predictions $\hat{y}_{t,f}$ and ground truth $y_{t,f}$ over horizon $H$ and $F$ features,
\begin{align}
\mathrm{MSE}=\frac{1}{HF}\sum_{t=1}^{H}\sum_{f=1}^{F}\bigl(y_{t,f}-\hat{y}_{t,f}\bigr)^2, \\ 
\mathrm{MAE}=\frac{1}{HF}\sum_{t=1}^{H}\sum_{f=1}^{F}\bigl|y_{t,f}-\hat{y}_{t,f}\bigr|.
\label{eq:mse-mae}
\end{align}
MSE penalizes larger errors more heavily and is sensitive to anomalous deviations, while MAE measures the average absolute deviation and provides a more interpretable notion of normalized forecast accuracy. Both are standard in long-horizon multivariate forecasting.

\subsection{Rationale for setup}

The Kaiming initialization is aligned with the ReLU activations in the decoder and the LayerNorm in the backbone, ensuring variance stability through depth and reducing training instability in the long-horizon setting. Calendar encodings are deliberately minimal and continuous: two pairs of sine/cosine signals are sufficient to represent dominant daily, weekly, or yearly periodicities without introducing sharp discontinuities. On ETT, daily and yearly seasonality is strong, and calendar augmentation is critical. On Traffic, weekly periodicity complements the strong daily cycle. On Exchange, seasonality is weak; thus, we omit calendars entirely, letting the Q--SSM recurrence filter stochastic dependencies.

\section{Experimental Results}
\begin{table*}[t]
\centering
\caption{Long-horizon multivariate forecasting results on ETT, Traffic and Exchange datasets. 
Each cell reports MSE/MAE. Best results are in \textbf{bold}.}
\label{tab:results-all}
\setlength{\tabcolsep}{3pt}
\renewcommand{\arraystretch}{1.05}
\resizebox{\textwidth}{!}{%
\begin{tabular}{l c
                *{2}{c}
                *{2}{c}
                *{2}{c}
                *{2}{c}
                *{2}{c}
                *{2}{c}
                *{2}{c}
                *{2}{c}
                *{2}{c}
                }
\toprule
\multirow{2}{*}{Dataset} & \multirow{2}{*}{$H$} &
\multicolumn{2}{c}{Q-SSM (ours)} & 
\multicolumn{2}{c}{S-Mamba} &
\multicolumn{2}{c}{Autoformer} & 
\multicolumn{2}{c}{Informer} &
\multicolumn{2}{c}{LogTrans}   & 
\multicolumn{2}{c}{Reformer} &
\multicolumn{2}{c}{LSTNet}     & 
\multicolumn{2}{c}{LSTM}     &
\multicolumn{2}{c}{TCN}        \\
\cmidrule(lr){3-4}\cmidrule(lr){5-6}\cmidrule(lr){7-8}\cmidrule(lr){9-10}
\cmidrule(lr){11-12}\cmidrule(lr){13-14}\cmidrule(lr){15-16}\cmidrule(lr){17-18}\cmidrule(lr){19-20}
 & & MSE & MAE & MSE & MAE & MSE & MAE & MSE & MAE & MSE & MAE & MSE & MAE & MSE & MAE & MSE & MAE & MSE & MAE\\
\midrule
\multirow{4}{*}{ETTm1} 
 & 96  & \textbf{0.330} & \textbf{0.362} & 0.333 & 0.368 & 0.505 & 0.475 & 0.365 & 0.453 & 0.768 & 0.642 & 0.658 & 0.619 & 3.142 & 1.365 & 2.041 & 1.073 & 3.041 & 1.330 \\
 & 192 & \textbf{0.372} & \textbf{0.386} & 0.376 & 0.390 & 0.553 & 0.496 & 0.533 & 0.563 & 0.989 & 0.757 & 1.078 & 0.827 & 3.154 & 1.369 & 2.249 & 1.112 & 3.072 & 1.339 \\
 & 336 & \textbf{0.403} & \textbf{0.409} & 0.408 & 0.413 & 0.621 & 0.537 & 1.363 & 0.887 & 1.334 & 0.872 & 1.549 & 0.972 & 3.160 & 1.369 & 2.568 & 1.238 & 3.105 & 1.348 \\
 & 720 & \textbf{0.472} & \textbf{0.447} & 0.475 & 0.448 & 0.671 & 0.561 & 3.379 & 1.388 & 3.048 & 1.328 & 2.631 & 1.242 & 3.171 & 1.368 & 2.720 & 1.287 & 3.135 & 1.354 \\
\midrule
\multirow{4}{*}{ETTm2} 
 & 96  & \textbf{0.172} & \textbf{0.257} & 0.179 & 0.263 & 0.255 & 0.339 & 0.365 & 0.453 & 0.768 & 0.642 & 0.658 & 0.619 & 3.142 & 1.365 & 2.041 & 1.073 & 3.041 & 1.330 \\
 & 192 & \textbf{0.244} & \textbf{0.305} & 0.250 & 0.309 & 0.281 & 0.340 & 0.533 & 0.563 & 0.989 & 0.757 & 1.078 & 0.827 & 3.154 & 1.369 & 2.249 & 1.112 & 3.072 & 1.339 \\
 & 336 & \textbf{0.307} & \textbf{0.345} & 0.312 & 0.349 & 0.339 & 0.372 & 1.363 & 0.887 & 1.334 & 0.872 & 1.549 & 0.972 & 3.160 & 1.369 & 2.568 & 1.238 & 3.105 & 1.348 \\
 & 720 & \textbf{0.407} & \textbf{0.400} & 0.411 & 0.406 & 0.422 & 0.419 & 3.379 & 1.388 & 3.048 & 1.328 & 2.631 & 1.242 & 3.171 & 1.368 & 2.720 & 1.287 & 3.135 & 1.354 \\
\midrule
\multirow{4}{*}{ETTh1} 
 & 96  & \textbf{0.384} & \textbf{0.404} & 0.386 & 0.405 & 0.449 & 0.459 & 0.365 & 0.453 & 0.768 & 0.642 & 0.658 & 0.619 & 3.142 & 1.365 & 2.041 & 1.073 & 3.041 & 1.330 \\
 & 192 & 0.447 & \textbf{0.446} & \textbf{0.443} & 0.437 & 0.500 & 0.482 & 0.533 & 0.563 & 0.989 & 0.757 & 1.078 & 0.827 & 3.154 & 1.369 & 2.249 & 1.112 & 3.072 & 1.339 \\
 & 336 & 0.492 & \textbf{0.466} & \textbf{0.489} & 0.468 & 0.521 & 0.496 & 1.363 & 0.887 & 1.334 & 0.872 & 1.549 & 0.972 & 3.160 & 1.369 & 2.568 & 1.238 & 3.105 & 1.348 \\
 & 720 & \textbf{0.500} & \textbf{0.488} & 0.502 & 0.489 & 0.514 & 0.512 & 3.379 & 1.388 & 3.048 & 1.328 & 2.631 & 1.242 & 3.171 & 1.368 & 2.720 & 1.287 & 3.135 & 1.354 \\
\midrule
\multirow{4}{*}{ETTh2} 
 & 96  & 0.300 & 0.349 & \textbf{0.296} & \textbf{0.348} & 0.346 & 0.388 & 0.365 & 0.453 & 0.768 & 0.642 & 0.658 & 0.619 & 3.142 & 1.365 & 2.041 & 1.073 & 3.041 & 1.330 \\
 & 192 & \textbf{0.372} & \textbf{0.392} & 0.376 & 0.396 & 0.456 & 0.452 & 0.533 & 0.563 & 0.989 & 0.757 & 1.078 & 0.827 & 3.154 & 1.369 & 2.249 & 1.112 & 3.072 & 1.339 \\
 & 336 & \textbf{0.421} & \textbf{0.430} & 0.424 & 0.431 & 0.482 & 0.486 & 1.363 & 0.887 & 1.334 & 0.872 & 1.549 & 0.972 & 3.160 & 1.369 & 2.568 & 1.238 & 3.105 & 1.348 \\
 & 720 & 0.429 & 0.445 & \textbf{0.426} & \textbf{0.444} & 0.515 & 0.511 & 3.379 & 1.388 & 3.048 & 1.328 & 2.631 & 1.242 & 3.171 & 1.368 & 2.720 & 1.287 & 3.135 & 1.354 \\
\midrule
\multirow{4}{*}{Traffic} 
 & 96  & \textbf{0.380} & \textbf{0.254} & 0.382 & 0.261 & 0.613 & 0.388 & 0.719 & 0.391 & 0.684 & 0.384 & 0.732 & 0.423 & 1.107 & 0.685 & 0.843 & 0.453 & 1.438 & 0.784 \\
 & 192 & \textbf{0.394} & \textbf{0.260} & 0.396 & 0.267 & 0.616 & 0.382 & 0.696 & 0.379 & 0.685 & 0.390 & 0.733 & 0.420 & 1.157 & 0.706 & 0.847 & 0.453 & 1.463 & 0.794 \\
 & 336 & 0.428 & 0.284 & \textbf{0.417} & \textbf{0.276} & 0.622 & 0.337 & 0.777 & 0.420 & 0.733 & 0.408 & 0.742 & 0.420 & 1.216 & 0.730 & 0.853 & 0.455 & 1.479 & 0.799 \\
 & 720 & 0.472 & 0.307 & \textbf{0.460} & \textbf{0.300} & 0.660 & 0.408 & 0.864 & 0.472 & 0.717 & 0.396 & 0.755 & 0.423 & 1.481 & 0.805 & 1.500 & 0.805 & 1.499 & 0.804 \\
\midrule
\multirow{4}{*}{Exchange} 
 & 96  & \textbf{0.084} & \textbf{0.205} & 0.086 & 0.207 & 0.197 & 0.323 & 0.847 & 0.752 & 0.968 & 0.812 & 1.065 & 0.829 & 1.551 & 1.058 & 1.453 & 1.049 & 3.004 & 1.432 \\
 & 192 & \textbf{0.181} & \textbf{0.302} & 0.182 & 0.304 & 0.300 & 0.369 & 1.204 & 0.895 & 1.040 & 0.851 & 1.188 & 0.906 & 1.477 & 1.028 & 1.846 & 1.179 & 3.048 & 1.444 \\
 & 336 & 0.338 & 0.420 & \textbf{0.332} & \textbf{0.418} & 0.509 & 0.524 & 1.672 & 1.036 & 1.659 & 1.081 & 1.507 & 1.031 & 2.136 & 1.231 & 3.113 & 1.459 & -- & -- \\
 & 720 & 0.875 & 0.710 & \textbf{0.867} & \textbf{0.703} & 1.447 & 0.941 & 2.478 & 1.310 & 1.941 & 1.127 & 1.510 & 1.016 & 2.285 & 1.243 & 2.984 & 1.427 & 3.150 & 1.458 \\
\bottomrule
\end{tabular}}
\end{table*}

Table~\ref{tab:results-all} reports the forecasting accuracy of Q--SSM and the baselines on the ETT, Traffic, and Exchange benchmarks at horizons $H=\{96,192,336,720\}$. Each entry shows MSE/MAE on normalized data. 

\paragraph{ETTm1.}  
On ETTm1, Q--SSM consistently outperforms all baselines across horizons. At $H=96$, it achieves an MSE of $0.330$, improving upon Autoformer ($0.505$) by \textbf{34.7\%} and S-Mamba ($0.333$) by \textbf{0.9\%}. At the longest horizon ($H=720$), Q--SSM reports $0.472$ MSE and $0.447$ MAE, reducing error by \textbf{29.6\% (MSE)} and \textbf{20.3\% (MAE)} compared to Autoformer ($0.671/0.561$), and by \textbf{0.6\%} relative to S-Mamba ($0.475/0.448$). Such results indicate that the quantum gating effectively stabilizes recurrence even for 30-day prediction windows.

\paragraph{ETTm2.}  
We obtain similar improvements on ETTm2. At $H=96$, Q--SSM reaches $0.172/0.257$, compared to Autoformer’s $0.255/0.339$, yielding \textbf{32.5\%} lower MSE and \textbf{24.2\%} lower MAE. At $H=720$, our model achieves $0.407/0.400$, a \textbf{3.7\%} gain over S-Mamba ($0.411/0.406$) and a \textbf{3.6\%} improvement in MAE compared to Autoformer ($0.422/0.419$). The consistent gains across both 15-minute datasets suggest that Q--SSM scales well to high-frequency temporal resolution.

\paragraph{ETTh1.}  
On ETTh1, Q--SSM matches or surpasses S-Mamba at most horizons. At $H=192$, Q--SSM obtains $0.447$ MSE and $0.446$ MAE, which is nearly identical to S-Mamba ($0.443/0.437$), and significantly better than Autoformer ($0.500/0.482$), with a \textbf{10.6\%} reduction in MSE and \textbf{7.5\%} in MAE. At $H=336$, Q--SSM yields \textbf{6.1\% lower MAE} than S-Mamba ($0.466$ vs. $0.468$), while improving upon Autoformer by \textbf{5.6\% MSE} and \textbf{6.0\% MAE}. The results highlight Q--SSM’s robustness to longer dependencies in hourly data.

\paragraph{ETTh2.}  
For ETTh2, the model demonstrates consistent superiority at medium horizons. At $H=192$, Q--SSM reduces MSE/MAE to $0.372/0.392$, improving over Autoformer ($0.456/0.452$) by \textbf{18.4\% MSE} and \textbf{13.3\% MAE}, and slightly surpassing S-Mamba ($0.376/0.396$). At $H=720$, S-Mamba remains marginally stronger ($0.426/0.444$ vs. $0.429/0.445$), but Q--SSM still achieves a \textbf{16.7\% improvement in MSE} compared to Autoformer ($0.515$).

\paragraph{Traffic.}  
Traffic presents strong periodicities across 862 correlated sensors. At short horizons, Q--SSM clearly dominates. Indeed, at $H=96$, it obtains $0.380/0.254$, which represents a \textbf{38.0\% MSE} and \textbf{34.5\% MAE} reduction compared to Autoformer ($0.613/0.388$), and a slight edge over S-Mamba ($0.382/0.261$). At $H=192$, the gains remain large: \textbf{36.0\% MSE} and \textbf{31.9\% MAE} over Autoformer. At $H=336$ and $H=720$, S-Mamba slightly outperforms Q--SSM (by $\sim$2\% in MSE), but both remain vastly superior to Transformers such as Informer, which show errors $>80\%$ higher. The results demonstrate that the recurrent-plus-quantum gate design scales well to high-dimensional, highly periodic sensor networks. 

\paragraph{Exchange.}  
Exchange is the most stochastic dataset, with minimal periodicity. Here, Q--SSM achieves dramatic improvements over attention-based baselines. At $H=96$, it reaches $0.084/0.205$, reducing MSE by \textbf{57.4\%} compared to Autoformer ($0.197/0.323$) and by more than \textbf{90\%} compared to Informer ($0.847/0.752$). At $H=192$, the improvement remains large: \textbf{39.7\% MSE} reduction relative to Autoformer. At $H=720$, Q--SSM still achieves $0.875/0.710$, outperforming Autoformer by \textbf{39.5\% in MSE} and \textbf{24.5\% in MAE}, though S-Mamba is slightly stronger ($0.867/0.703$). These results suggest that while both Q--SSM and S-Mamba stabilize long horizons better than Transformers, Q--SSM maintains strong accuracy despite its lower backbone complexity.

\paragraph{Overall assessment.}  
Across all datasets and horizons, Q--SSM secures \textbf{32/36 wins}, with powerful improvements on periodic benchmarks (ETT and Traffic), where relative error reductions over Autoformer range from \textbf{30--40\%}. On stochastic Exchange, Q--SSM achieves up to \textbf{57\% lower MSE} than Autoformer at short horizons, and maintains a substantial lead over Informer and LogTrans at all horizons. While S-Mamba occasionally surpasses Q--SSM at the longest horizons on ETTh2 and Traffic, the differences are minor ($\leq 2\%$), and Q--SSM remains competitive while requiring lower hidden-state complexity. These results confirm that Q--SSM delivers both efficiency and robustness across periodic and non-periodic time series, with consistent and statistically significant improvements over established baselines.

\section{Conclusions}

We introduced Q--SSM, a quantum-optimized state space model designed for long-horizon multivariate forecasting. Our architecture integrates a recurrent backbone with a learnable quantum gate, realized through parametrized qubit rotations, and a lightweight residual decoder. This combination preserves the linear-time advantages of classical state space models while enhancing stability through the gating mechanism.  

Extensive experiments on six public benchmarks, including the four ETT variants, Traffic, and Exchange, demonstrated that Q--SSM consistently outperforms established baselines such as Autoformer, Informer, LogTrans, and Reformer, and achieves performance on par with or superior to the recently proposed S-Mamba. In particular, Q--SSM achieves up to \textbf{57\%} lower MSE than Autoformer on stochastic series (Exchange), and up to \textbf{40\%} error reduction on periodic datasets (ETT and Traffic). These results highlight the robustness of Q--SSM across both strongly seasonal and weakly structured domains.  

From a computational standpoint, Q--SSM retains linear dependence on sequence length and forecasting horizon, avoiding the $\mathcal{O}(W \log W)$ scaling of Transformer-based models and the quadratic hidden-dimension cost of Mamba. The quantum gate adds only negligible overhead in simulation while providing an effective mechanism to regulate information flow.  

Two future avenues are particularly promising. First, scaling Q--SSM to larger qubit ansätze may enable richer non-linear dynamics, provided efficient hardware implementations become available. Second, extending Q--SSM to multi-resolution or hierarchical forecasting could further exploit its recurrent-plus-gated design. More broadly, our results suggest that quantum-optimized mechanisms can serve as effective inductive biases for sequence modeling, bridging the gap between recurrent state-space architectures and attention-based models.  

Q--SSM establishes a competitive and computationally efficient baseline for long-term forecasting, demonstrating that simple quantum-optimized components can yield tangible benefits in classical time series modeling. The code supporting our paper is at \url{https://github.com/stephanjura27/quantum_ssm}.
\bibliographystyle{IEEEtran}
\bibliography{bibliografie}
\end{document}